# A Hierarchy of Tractable Subsets
# for Computing Stable Models

**Rachel Ben-Eliyahu**                                        RACHEL@CS.BGU.AC.IL
*Mathematics and Computer Science Department*
*Ben-Gurion University of the Negev*
*P.O.B. 653, Beer-Sheva 84105, Israel*

## Abstract

Finding the stable models of a knowledge base is a significant computational problem in artificial intelligence. This task is at the computational heart of truth maintenance systems, autoepistemic logic, and default logic. Unfortunately, it is NP-hard. In this paper we present a hierarchy of classes of knowledge bases, $\Omega_1, \Omega_2, ...$, with the following properties: first, $\Omega_1$ is the class of all stratified knowledge bases; second, if a knowledge base $\Pi$ is in $\Omega_k$, then $\Pi$ has at most $k$ stable models, and all of them may be found in time $O(lnk)$, where $l$ is the length of the knowledge base and $n$ the number of atoms in $\Pi$; third, for an arbitrary knowledge base $\Pi$, we can find the minimum $k$ such that $\Pi$ belongs to $\Omega_k$ in time polynomial in the size of $\Pi$; and, last, where $\mathcal{K}$ is the class of all knowledge bases, it is the case that $\bigcup_{i=1}^{\infty} \Omega_i = \mathcal{K}$, that is, every knowledge base belongs to some class in the hierarchy.

## 1. Introduction

The task of computing the stable models of a knowledge base lies at the heart of three of the fundamental systems in Artificial Intelligence (AI): truth maintenance systems (TMSs), default logic, and autoepistemic logic. Yet, this task is intractable (Elkan, 1990; Kautz & Selman, 1991; Marek & Truszczyński, 1991). In this paper, we introduce a hierarchy of classes of knowledge bases which achieves this task in polynomial time. Membership in a certain class in the hierarchy is testable in polynomial time. Hence, given a knowledge base, the cost of computing its stable models can be bounded prior to the actual computation (if the algorithms on which this hierarchy is based are used).

First, let us elaborate the relevance of computing stable models to AI tasks. We define a knowledge base to be a set of rules of the form

$$C \longleftarrow A_1, ..., A_m, not\ B_1, ..., not\ B_n \tag{1}$$

where $C$, all $A$s, and all $B$s are atoms in some propositional language. Substantial efforts to give a meaning, or semantics, to a knowledge base have been made in the logic programming community (Przymusinska & Przymusinski, 1990). One of the most successful semantics for knowledge bases is *stable model semantics* (Bidoit & Froidevaux, 1987; Gelfond & Lifschitz, 1988; Fine, 1989), which associates any knowledge base with a (possibly empty) set of models called *stable models*. Intuitively, each stable model represents a set of coherent conclusions one might deduce from the knowledge base. It turns out that stable models play a central role in some major deductive systems in AI. [1]

## 1.1 Stable Models and TMSs

TMSs (Doyle, 1979) are inference systems for nonmonotonic reasoning with default assumptions. The TMS manages a set of nodes and a set of justifications, where each node represents a piece of information and the justifications are rules that state the dependencies between the nodes. The TMS computes a *grounded* set of nodes and assigns this set to be the information believed to be true at a given point in time. Intuitively, a set of believed nodes is grounded if it satisfies all the rules, but no node is believed true solely on the basis of a circular chain of justifications. Elkan (1990) pointed out that the nodes of a TMS can be viewed as propositional atoms, and the set of its justifications as a knowledge base. He showed that the task of computing grounded interpretations for a set of TMS justifications corresponds exactly to the task of computing the stable models of the knowledge base represented by the set of TMS justifications.

## 1.2 Stable Models and Autoepistemic Logic

Autoepistemic logic was invented by Moore (1985) in order to formalize the process of an agent reasoning about its own beliefs. The language of autoepistemic logic is a propositional language augmented by a modal operator $\mathbf{L}$. Given a theory (a set of formulas) $T$ in autoepistemic logic, a theory $E$ is called a *stable expansion* of $T$ iff

$$E = (T \bigcup \{\mathbf{L} F | F \in E\} \bigcup \{\neg \mathbf{L} F | F \notin E\})^*$$

where $T^*$ denotes the logical closure of $T$. We will now restrict ourselves to a subset of autoepistemic logic in which each formula is of the form

$$A_1 \wedge ... \wedge A_m \wedge \neg \mathbf{L} B_1 \wedge ... \wedge \neg \mathbf{L} B_n \longrightarrow C \tag{2}$$

where $C$, each of the $A$s, and each of the $B$s are propositional atoms. We call this subset the class of *autoepistemic programs*. Every autoepistemic program $T$ can be translated into a knowledge base $\Pi_T$ by representing the formula (2) as the knowledge base rule (1). Elkan (1990) has shown that $M$ is a stable model of $\Pi_T$ iff there is an expansion $E$ of $T$ such that $M$ is the set of all positive atoms in $E$. Thus, algorithms for computing stable models may be used in computing expansions of autoepistemic programs. The relationship between stable model semantics and autoepistemic logic has also been explored by Gelfond (1987) and Gelfond and Lifschitz (1988, 1991).

## 1.3 Stable Models and Default Logic

Default logic is a formalism developed by Reiter (1980) for reasoning with default assumptions. A default theory can be viewed as a set of defaults, and a *default* is defined as an expression of the form

$$\frac{\alpha : \beta_1, ..., \beta_n}{\gamma} \tag{3}$$

---

1. In logic programming terminology, the knowledge bases discussed in this paper are called *normal logic programs*.

where $\alpha$, $\gamma$, and $\beta_1, ..., \beta_n$ are formulas in some first-order language. According to Reiter, $E$ is an extension for a default theory $\Delta$ iff $E$ coincides with one of the minimal deductively closed sets of sentences $E'$ satisfying the condition[2] that for any grounded instance of a default (3) from $\Delta$, if $\alpha \in E'$ and $\neg\beta_1, ..., \neg\beta_n \notin E$, then $\gamma \in E'$.

Now consider the subset of default theories that we call *default programs*. A *default program* is a set of defaults of the form

$$\frac{A_1 \wedge ... \wedge A_m : \neg B_1, ..., \neg B_n}{C} \tag{4}$$

in which $C$, each of the $A$s, and each of the $B$s are atoms in a propositional language.

Each default program $\Delta$ can be associated with a knowledge base $\Pi_\Delta$ by replacing each default of the form (4) with the rule (1).

Gelfond and Lifschitz (1991) have shown that the logical closure of a set of atoms $E$ is an extension of $\Delta$ iff $E$ is a stable model of $\Pi_\Delta$. Algorithms for computing stable models can thus be used in computing extensions of Reiter's default theories.

$$* \; * \; *$$

The paper is organized as follows. In the next section, we define our terminology. Section 3 presents two algorithms for computing all stable models of a knowledge base. The complexity of the first of these algorithms depends on the number of atoms appearing negatively in the knowledge base, while the complexity of the other algorithm depends on the number of rules having negative atoms in their bodies. In Section 4, we present the main algorithm of the paper, called *algorithm AAS*. Algorithm AAS works from the bottom up on the superstructure of the dependency graph of the knowledge base and uses the two algorithms presented in Section 3 as subroutines. Section 5 explains how the AAS algorithm can be generalized to handle knowledge bases over a first-order language. Finally, in Sections 6 and 7, we discuss related work and make concluding remarks.

## 2. Preliminary Definitions

Recall that here a knowledge base is defined as a set of rules of the form

$$C \longleftarrow A_1, ..., A_m, \textit{not } B_1, ..., \textit{not } B_n \tag{5}$$

where $C$, each of the $A$s, and each of the $B$s are propositional atoms. The expression to the left of $\longleftarrow$ is called the *head* of the rule, while the expression to the right of $\longleftarrow$ is called the *body* of the rule. Each of the $A$s is said to *appear positive* in the rule, and, accordingly, each of the $B$s is said to *appear negative* in the rule. Rule (5) is said to be *about* $C$. A rule with an empty body is called a *unit rule*. Sometimes we will treat a truth assignment (in other words, interpretation) in propositional logic as a set of atoms — the set of all atoms assigned **true** by the interpretation. Given an interpretation $I$ and a set of atoms $A$, $I_A$ denotes the projection of $I$ over $A$. Given two interpretations, $I$ and $J$, over sets of atoms

---

2. Note the appearance of $E$ in the condition.

$A$ and $B$, respectively, the interpretation $I + J$ is defined as follows:

$$I + J(P) = \begin{cases} I(P) & \text{if } P \in A \setminus B \\ J(P) & \text{if } P \in B \setminus A \\ I(P) & \text{if } P \in A \bigcap B \text{ and } I(P) = J(P) \\ \text{undefined} & \text{otherwise} \end{cases}$$

If $I(P) = J(P)$ for every $P \in A \bigcap B$, we say that $I$ and $J$ are *consistent*.

A *partial* interpretation is a truth assignment over a subset of the atoms. Hence, a partial interpretation can be represented as a consistent set of literals: positive literals represent the atoms that are true, negative literals the atoms that are false, and the rest are unknown. A knowledge base will be called *Horn* if all its rules are Horn. A model for a theory (set of clauses) in propositional logic is a truth assignment that satisfies all the clauses. If one looks at a knowledge base as a theory in propositional logic, a Horn knowledge base has a unique minimal model (recall that a model $m$ is minimal among a set of models $M$ iff there is no model $m' \in M$ such that $m' \subset m$).

Given a knowledge base $\Pi$ and a set of atoms $m$, Gelfond and Lifschitz defined what is now called the *Gelfond-Lifschitz (GL) transform* of $\Pi$ w.r.t. $m$, which is a knowledge base $\Pi_m$ obtained from $\Pi$ by deleting each rule that has a negative literal *not P* in its body with $P \in m$ and deleting all negative literals in the bodies of the remaining rules. Note that $\Pi_m$ is a Horn knowledge base. A model $m$ is a *stable* model of a knowledge base $\Pi$ iff it is the unique minimal model of $\Pi_m$ (Gelfond & Lifschitz, 1988).

**Example 2.1** Consider the following knowledge base $\Pi_0$, which will be used as one of the canonical examples throughout this paper:

$$
\begin{aligned}
warm\_blooded &\longleftarrow mammal & (6) \\
live\_on\_land &\longleftarrow mammal, not\, ab1 & (7) \\
female &\longleftarrow mammal, not\, male & (8) \\
male &\longleftarrow mammal, not\, female & (9) \\
mammal &\longleftarrow dolphin & (10) \\
ab1 &\longleftarrow dolphin & (11) \\
mammal &\longleftarrow lion & (12) \\
lion &\longleftarrow & (13)
\end{aligned}
$$

$m = \{lion, mammal, warm\_blooded, live\_on\_land, female\}$ is a stable model of $\Pi_0$. Indeed, $\Pi_{0_m}$ (the GL transform of $\Pi_0$ w.r.t. $m$) is

$$
\begin{aligned}
warm\_blooded &\longleftarrow mammal \\
live\_on\_land &\longleftarrow mammal \\
female &\longleftarrow mammal \\
mammal &\longleftarrow dolphin \\
ab1 &\longleftarrow dolphin \\
mammal &\longleftarrow lion \\
lion &\longleftarrow
\end{aligned}
$$

and $m$ is a minimal model of $\Pi_{0_m}$.

A set of atoms $S$ *satisfies the body* of a rule $\delta$ iff each atom that appears positive in the body of $\delta$ is in $S$ and each atom that appears negative in the body of $\delta$ is not in $S$. A set of atoms $S$ *satisfies a rule* iff either it does not satisfy its body, or it satisfies its body and the atom that appears in its head belongs to $S$.

A *proof* of an atom is a sequence of rules from which the atom can be derived. Formally, we can recursively define when an atom $P$ has a proof w.r.t. a set of atoms $S$ and a knowledge base $\Pi$:

- If the unit rule $P \longleftarrow$ is in $\Pi$, then $P$ has a proof w.r.t. $\Pi$ and $S$.

- If the rule $P \longleftarrow A_1, ..., A_m, not\ B_1, ..., not\ B_n$ is in $\Pi$, and for every $i = 1, ..., n$ $B_i$ is not in $S$, and for every $i = 1, ..., m$ $A_i$ already has a proof w.r.t. $\Pi$ and $S$, then $P$ has a proof w.r.t. $\Pi$ and $S$.

**Theorem 2.2** (Elkan, 1990; Ben-Eliyahu & Dechter, 1994) *A set of atoms $S$ is a stable model of a knowledge base $\Pi$ iff*

*1. $S$ satisfies each rule in $\Pi$, and*

*2. for each atom $P$ in $S$, there is a proof of $P$ w.r.t $\Pi$ and $S$.*

It is a simple matter to show that the following lemma is true.

**Lemma 2.3** *Let $\Pi$ be a knowledge base, and let $S$ be a set of atoms. Define:*

*1. $S_0 = \emptyset$, and*

*2. $S_{i+1} = S_i \bigcup \{P \,|\, P \longleftarrow A_1, ..., A_m, not\ B_1, ..., not\ B_n$ is in $\Pi$,
all of the $A$'s belong to $S_i$ and none of the $B$'s belong to $S\}$.*

*Then $S$ is a stable model of $\Pi$ iff $S = \bigcup_0^\infty S_i$.*

Observe that although every stable model is a minimal model of the knowledge base viewed as a propositional theory, not every minimal model is a stable model.

**Example 2.4** Consider the knowledge base

$$b \quad \longleftarrow \quad not\ a$$

Both $\{a\}$ and $\{b\}$ are minimal models of the knowledge base above, but only $\{b\}$ is a stable model of this knowledge base.

Note that a knowledge base may have one or more stable models, or no stable model at all. If a knowledge base has at least one stable model, we say that it is *consistent*.

The *dependency graph* of a knowledge base $\Pi$ is a directed graph where each atom is a node and where there is a *positive edge* directed from $P$ to $Q$ iff there is a rule about $Q$ in $\Pi$ in which $P$ appears positive in the body. Accordingly, there is a *negative edge* from $P$ to $Q$ iff there is a rule about $Q$ in which $P$ appears negative in the body. Recall that a *source* of a directed graph is a node with no incoming edges, while a *sink* is a node with no outgoing edges. Given a directed graph $G$ and a node $s$ in $G$, the *subgraph rooted by $s$* is the subgraph of $G$ having only nodes $t$ such that there is a path directed from $t$ to $s$ in $G$. The *children* of $s$ in $G$ are all nodes $t$ such that there is an arc directed from $t$ to $s$ in $G$.

**Example 2.5** The dependency graph of $\Pi_0$ is shown in Figure 1. Negative edges are marked "not." The children of *mammal* are *lion* and *dolphin*. The subgraph rooted by *on_land* is the subgraph that include the nodes *lion*, *mammal*, *dolphin*, *ab1*, and *on_land*.

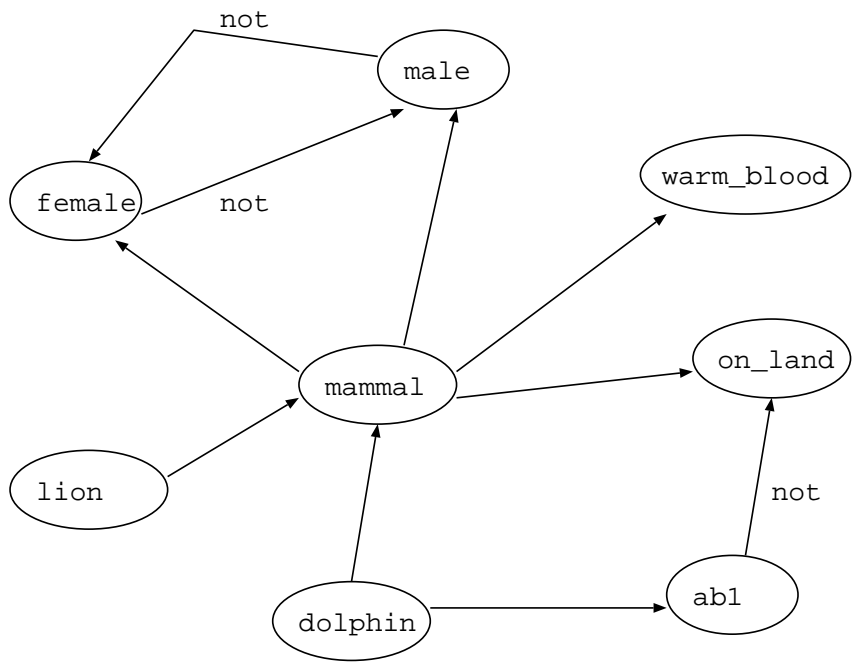

Figure 1: The dependency graph of $\Pi_0$

A knowledge base is *stratified* iff we can assign each atom $C$ a positive integer $i_C$ such that for every rule in the form of (5) above, for each of the $A$s, $i_A \leq i_C$, and for each of the $B$s, $i_B < i_C$. It can be readily demonstrated that a knowledge base is stratified iff in its dependency graph there are no directed cycles going through negative edges. It is well known in the logic programming community that a stratified knowledge base has a unique stable model that can be found in linear time (Gelfond & Lifschitz, 1988; Apt, Blair, & Walker, 1988).

**Example 2.6** $\Pi_0$ is not a stratified knowledge base. The following knowledge base, $\Pi_1$, is stratified (we can assign *ab2* and *penguin* the number 1, and each of the other atoms the number 2):

$$
\begin{aligned}
live\_on\_land &\longleftarrow bird \\
fly &\longleftarrow bird, not\ ab2 \\
bird &\longleftarrow penguin \\
ab2 &\longleftarrow penguin
\end{aligned}
$$

The strongly connected components of a directed graph $G$ make up a partition of its set of nodes such that, for each subset $S$ in the partition and for each $x, y \in S$, there are directed paths from $x$ to $y$ and from $y$ to $x$ in $G$. The strongly connected components are identifiable in linear time (Tarjan, 1972).

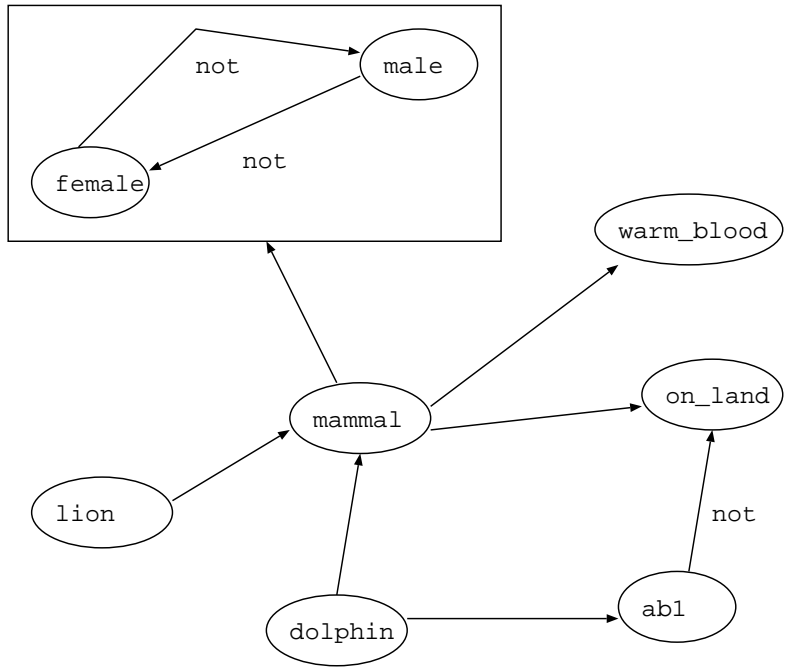

Figure 2: The super dependency graph of $\Pi_0$

The *super dependency graph* of a knowledge base $\Pi$, denoted $G_\Pi$, is the superstructure of the dependency graph of $\Pi$. That is, $G_\Pi$ is a directed graph built by making each strongly connected component in the dependency graph of $\Pi$ into a node in $G_\Pi$. An arc exists from a node $s$ to a node $v$ iff there is an arc from one of the atoms in $s$ to one of the atoms in $v$ in the dependency graph of $\Pi$. Note that $G_\Pi$ is an acyclic graph.

**Example 2.7** The super dependency graph of $\Pi_0$ is shown in Figure 2. The nodes in the square are grouped into a single node.

## 3. Two Algorithms for Computing Stable Models

The main contribution of this paper is the presentation of an algorithm whose efficiency depends on the "distance" of the knowledge base from a stratified knowledge base. This distance will be measured precisely in Section 4. We will first describe two other algorithms for computing stable models. These two algorithms do not take into account the level of "stratifiability" of the knowledge base, that is, they will still work in exponential time for stratified knowledge bases. Our main algorithm will use these two algorithms as procedures.

Given a truth assignment for a knowledge base, we can verify in polynomial time whether it is a stable model by using Lemma 2.3. Therefore, a straightforward algorithm for computing all stable models can simply check all possible truth assignments and determine whether each of them is a stable model. The time complexity of this straightforward procedure will be exponential in the number of atoms used in the knowledge base. Below, we present two algorithms that can often function more efficiently than the straightforward procedure.

## 3.1 An Algorithm That Depends on the Number of Negative Atoms in the Knowledge Base

Algorithm ALL-STABLE1 (Figure 3) enables us to find all the stable models in time exponential in the number of the atoms that appear negative in the knowledge base.

The algorithm follows from work on abductive extensions of logic programming in which stable models are characterized in terms of sets of hypotheses that can be drawn as additional information (Eshghi & Kowalski, 1989; Dung, 1991; Kakas & Mancarella, 1991). This is done by making negative atoms abductible and by imposing appropriate denials and disjunctions as integrity constraints. The work of Eshghi and Kowalski (1989), Dung (1991), and Kakas and Mancarella (1991) implies the following.

**Theorem 3.1** *Let $\Pi$ be a knowledge base, and let $H$ be the set of atoms that appear negated in $\Pi$. $M$ is a stable model of $\Pi$ iff there is an interpretation $I$ over $H$ such that*

*1. for every atom $P \in H$, if $P \in I$, then $P \in M'$,*

*2. $M'$ and $I$ are consistent, and*

*3. $M = I + M'$,*

*where $M'$ is the unique stable model of $\Pi_I$.*

**Proof:** The proof follows directly from the definition of stable models. Suppose $M$ is a stable model of a knowledge base $\Pi$, and let $H$ be the set of atoms that appear negative in $\Pi$. Then, by definition, $M$ is a stable model of $\Pi_M$. But note that $\Pi_M = \Pi_{M_H}$. Hence, the conditions of Theorem 3.1 hold for $M$, taking $M' = M$ and $I = M_H$. Now, suppose $\Pi$ is a knowledge base and $M = M' + I$, where $M'$ and $I$ are as in Theorem 3.1. Observe that $\Pi_M = \Pi_I$ and, hence, since $M'$ is a stable model of $\Pi_I$, $M'$ is a stable model of $\Pi_M$. We will show that $M$ is a stable model of $\Pi_M$. First, note that by condition 1, $M \subseteq M'$. Thus, $M$ satisfies all the rules in $\Pi_M$ and, if an atom $P$ has a proof w. r. t. $M'$ and $\Pi_M$, it has also a proof w. r. t. $M$ and $\Pi_M$. So, by Theorem 2.2, $M$ is a stable model of $\Pi_M$ and, by definition, $M$ is a stable model of $\Pi$. □

Theorem 3.1 implies algorithm ALL-STABLE1 (Figure 3), which computes all stable models of a knowledge base $\Pi$. Hence, we have the following complexity analysis.

**Proposition 3.2** *A knowledge base in which at most $k$ atoms appear negated has at most $2^k$ stable models and all of them can be found in time $O(nl2^k)$, where $l$ is the size of the knowledge base and $n$ the number of atoms used in the knowledge base.*

**Proof:** Follows from the fact that computing $\Pi_I$ and computing the unique stable model of a positive knowledge base is $O(nl)$. □

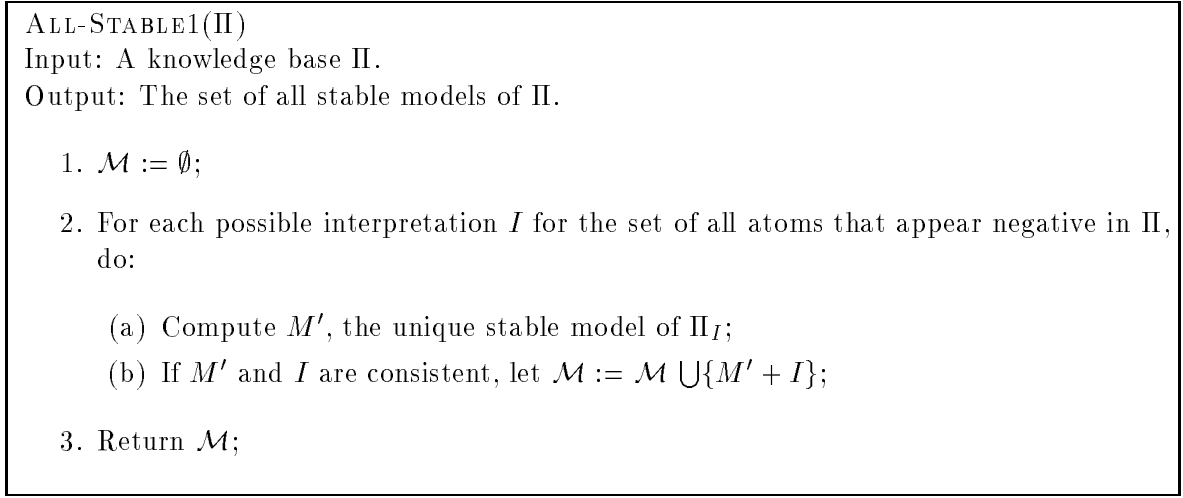

Figure 3: Algorithm ALL-STABLE1

## 3.2 An Algorithm That Depends on the Number of Non-Horn Rules

Algorithm ALL-STABLE2 (Figure 4) depends on the number of rules in which there are negated atoms. It gets as input a knowledge base $\Pi$, and, it outputs the set of all stable models of $\Pi$. This algorithm is based upon the observation that a stable model can be built by attempting all possible means of satisfying the negated atoms in bodies of non-Horn rules. Two procedures are called by ALL-STABLE2: UnitInst, shown in Figure 5; and NegUnitInst, shown in Figure 6. Procedure UnitInst gets as input a knowledge base $\Pi$ and a partial interpretation $m$. UnitInst looks recursively for unit rules in $\Pi$. For each unit rule $P \longleftarrow$, if $P$ is assigned **false** in $m$, it follows that $m$ cannot be a part of a model for $\Pi$, and the procedure returns **false**. If $P$ is not **false** in $m$, the procedure instantiates $P$ to **true** in the interpretation $m$ and deletes the positive appearances of $P$ from the body of each rule. It also deletes from $\Pi$ all the rules about $P$ and all the rules in which $P$ appears negative.

Procedure NegUnitInst receives as input a knowledge base $\Pi$, a partial interpretation $m$, and a set of atoms $Neg$. It first instantiates each atom in $Neg$ to **false** and then updates the knowledge base to reflect this instantiation. All the instantiations are recorded in $m$. In case of a conflict, namely, where the procedure tries to instantiate to **true** an atom that is already set to **false**, the procedure returns **false**; otherwise, it returns **true**.

**Proposition 3.3** *Algorithm* ALL-STABLE2 *is correct, that is, $m$ is a stable model of a knowledge base $\Pi$ iff it is generated by* ALL-STABLE2($\Pi$).

**Proof:** Suppose $m$ is a stable model of a knowledge base $\Pi$. Then, by Theorem 2.2, every atom set to **true** in $m$ has a proof w. r. t. $m$ and $\Pi$. Let $S$ be the set of all non-Horn rules whose bodies are satisfied by $m$. Clearly, at some point this $S$ is checked at step 3 of algorithm ALL-STABLE2. When this happens, all atoms that have a proof w. r. t. $m$ and $\Pi$ will be set to **true** by the procedure NegUnitInst (as can be proved by induction on the length of the proof). Hence, $m$ will be generated.

Suppose $m$ is generated by ALL-STABLE2($\Pi$). Obviously, every rule in $\Pi$ is satisfied by $m$ (step 3.c.ii), and every atom set to **true** by NegUnitInst has a proof w. r. t. $m$ and $\Pi$

---

ALL-STABLE2($\Pi$)

Input: A knowledge base $\Pi$.
Output: The set of all stable models of $\Pi$.

1. $M := \emptyset$;

2. Let $\Delta$ be the set of all non-Horn rules in $\Pi$.

3. For each subset $S$ of $\Delta$, do:

   (a) $Neg = \{P | not\ P$ is in the body of some rule in $S\}$;

   (b) $\Pi' := \Pi$; $m := \emptyset$;

   (c) If NegUnitInst($\Pi'$, $Neg$, $m$), then

      i. For each $P$ such that $m[P] = null$, let $m[P] := $ **false**;

      ii. If $m$ satisfies all the rules in $\Pi$, then $M := M \bigcup \{m\}$;

4. EndFor;

5. Return $M$;

---

Figure 4: Algorithm ALL-STABLE2

---

UnitInst($\Pi$, $m$)
Input: A knowledge base $\Pi$ and a partial interpretation $m$.
Output: Updates $m$ using the unit rules of $\Pi$. Returns **false** if there is a conflict between a unit rule and the value assigned to some atom in $m$; otherwise, returns **true** .

1. While $\Pi$ has unit rules, do:

   (a) Let $P \longleftarrow$ be a unit rule in $\Pi$;

   (b) If $m[P] = $ **false**, return **false**;

   (c) $m[P] := $ **true**;

   (d) Erase $P$ from the body of each rule in $\Pi$;

   (e) Erase from $\Pi$ all rules about $P$;

   (f) Erase from $\Pi$ all rules in which $P$ appears negative;

2. EndWhile;

3. Return **true**;

---

Figure 5: Procedure UnitInst

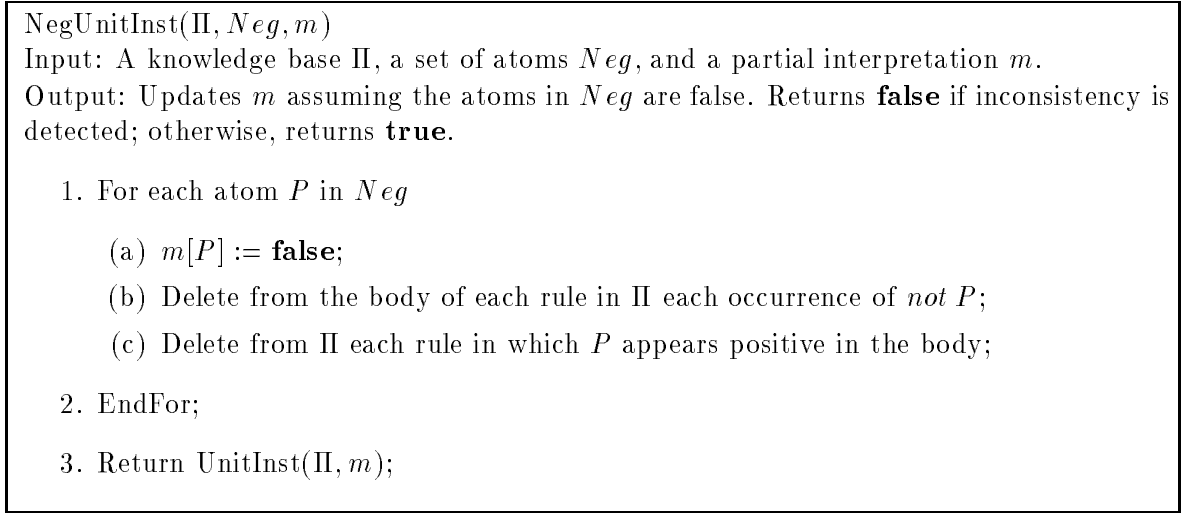

Figure 6: Procedure NegUnitInst

| | lion | dolphin | ab1 | mammal | warm_b | on_land | male | female |
|---|---|---|---|---|---|---|---|---|
| $S1$ | $T$ | $F$ | $F$ | $T$ | $T$ | $T$ | $F$ | $T$ |
| $S2$ | $T$ | $F$ | $F$ | $T$ | $T$ | $T$ | $T$ | $F$ |

Table 1: Models generated by Algorithm ALL-STABLE2

(as is readily observable from the way NegUnitInst works). Hence, by Theorem 2.2, $m$ is a stable model of $\Pi$. □

**Proposition 3.4** *A knowledge base having $c$ non-Horn rules has at most $2^c$ stable models and all of them can be found in time $O(nl2^c)$, where $l$ is the size of the knowledge base and $n$ the number of atoms used in the knowledge base.*

**Proof:** Straightforward, by induction on $c$. □

**Example 3.5** Suppose we call ALL-STABLE2 with $\Pi_0$ as the input knowledge base. At step 2, $\Delta$ is the set of rules (7), (8), and (9). When subsets of $\Delta$ which include both rules (8) and (9) are considered at step 3, NegUnitInst will return **false** because UnitInst will detect inconsistency. When the subset containing both rules (7) and (8) is considered, the stable model $S1$ of Table 1 will be generated. When the subset containing both rules (7) and (9) is considered, the stable model $S2$ of Table 1 will be generated. When all the other subsets that do not contain both rules (8) and (9) are tested at step 3, the $m$ generated will not satisfy all the rules in $\Pi$ and, hence, will not appear in the output.

Algorithms ALL-STABLE1 and ALL-STABLE2 do not take into account the structure of the knowledge base. For example, they are not polynomial for the class of stratified knowledge bases. We present next an algorithm that exploits the structure of the knowledge base.

## 4. A Hierarchy of Tractable Subsets Based on the Level of Stratifiability of the Knowledge Base

Algorithm Acyclic-All-Stable (AAS) in Figure 7 exploits the structure of the knowledge base as it is reflected in the super dependency graph of the knowledge base. It computes all stable models while traversing the super dependency graph from the bottom up, using the algorithms for computing stable models presented in the previous section as subroutines.

Let $\Pi$ be a knowledge base. With each node $s$ in $G_\Pi$ (the super dependency graph of $\Pi$), we associate $\Pi_s$, $A_s$, and $M_s$. $\Pi_s$ is the subset of $\Pi$ containing all the rules about the atoms in $s$, $A_s$ is the set of all atoms in the subgraph of $G_\Pi$ rooted by $s$, and $M_s$ is the set of stable models associated with the subset of the knowledge base $\Pi$ which contains only rules about atoms in $A_s$. Initially, $M_s$ is empty for every $s$. The algorithm traverses $G_\Pi$ from the bottom up. When at a node $s$, it first combines all the submodels of the children of $s$ into a single set of models $M_{c(s)}$. If $s$ is a source, then $M_{c(s)}$ is set to $\{\emptyset\}$[3]. Next, for each model $m$ in $M_{c(s)}$, AAS converts $\Pi_s$ to a knowledge base $\Pi_{s_m}$ using the GL transform and other transformations that depend on the atoms in $m$; then, it finds all the stable models of $\Pi_{s_m}$ and combines them with $m$. The set $M_s$ is obtained by repeating this operation for each $m$ in $M_{c(s)}$. AAS uses the procedure CartesProd (Figure 8), which receives as input several sets of models and returns the consistent portion of their Cartesian product. If one of the sets of models which CartesProd gets as input is the empty set, CartesProd will output an empty set of models. The procedure Convert gets as input a knowledge base $\Pi$, a model $m$, and a set of atoms $s$, and performs the following: for each atom $P$ in $m$, each positive occurrence of $P$ is deleted from the body of each rule in $\Pi$; for each rule in $\Pi$, if $not\ P$ is in the body of the rule and $P \in m$, then the rule is deleted from $\Pi$; if $not\ P$ is in the body of a rule and $P \notin m$, then, if $P \notin s$, $not\ P$ is deleted from that body. The procedure All-Stable called by AAS may be one of the procedures previously presented (All-Stable1 or All-Stable2) or it may be any other procedure that generates all stable models.

**Example 4.1** Suppose AAS is called to compute the stable models of $\Pi_0$. Suppose further that the algorithm traverses the super dependency graph in Figure 2 in the order $\{lion, dolphin, mammal, ab1, on\_land, warm\_blooded, female\text{-}male\}$ (recall that all the nodes inside the square make up one node that we are calling female-male or, for short, FM). After visiting all the nodes except the last, we have $M_{lion} = \{\{lion\}\}$, $M_{dolphin} = \{\emptyset\}$, $M_{mammal} = \{\{lion, mammal\}\}$, $M_{on\_land} = \{\{lion, mammal, onland\}\}$, $M_{warm\_blooded} = \{\{lion, mammal, warm\_blooded\}\}$. When visiting the node FM, we have after step 1.c that $M_{c(FM)} = M_{mammal}$. So step 1.d loops only once, for $m = \{lion, mammal\}$. Recall that $\Pi_{FM}$ is the knowledge base

$$female \quad \longleftarrow \quad mammal, not\ male$$
$$male \quad \longleftarrow \quad mammal, not\ female$$

---

3. Note the difference between $\{\emptyset\}$, which is a set of one model - the model that assigns **false** to all the atoms, and $\emptyset$, which is a set that contains no models.

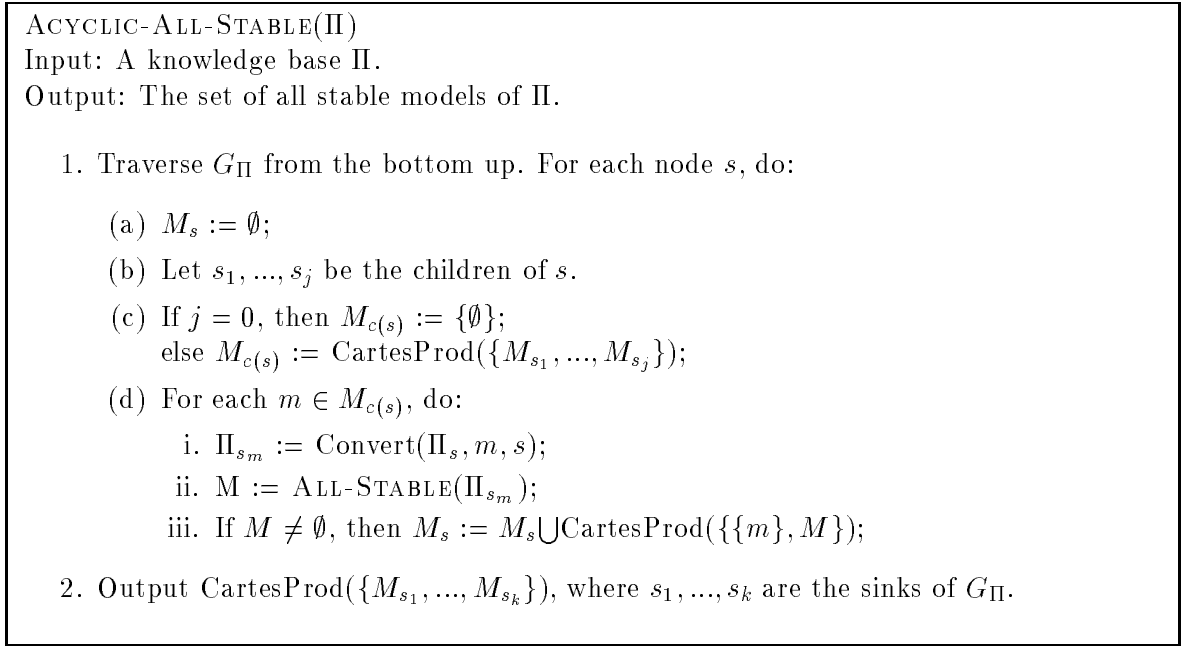

Figure 7: Algorithm Acyclic-All-Stable (AAS)

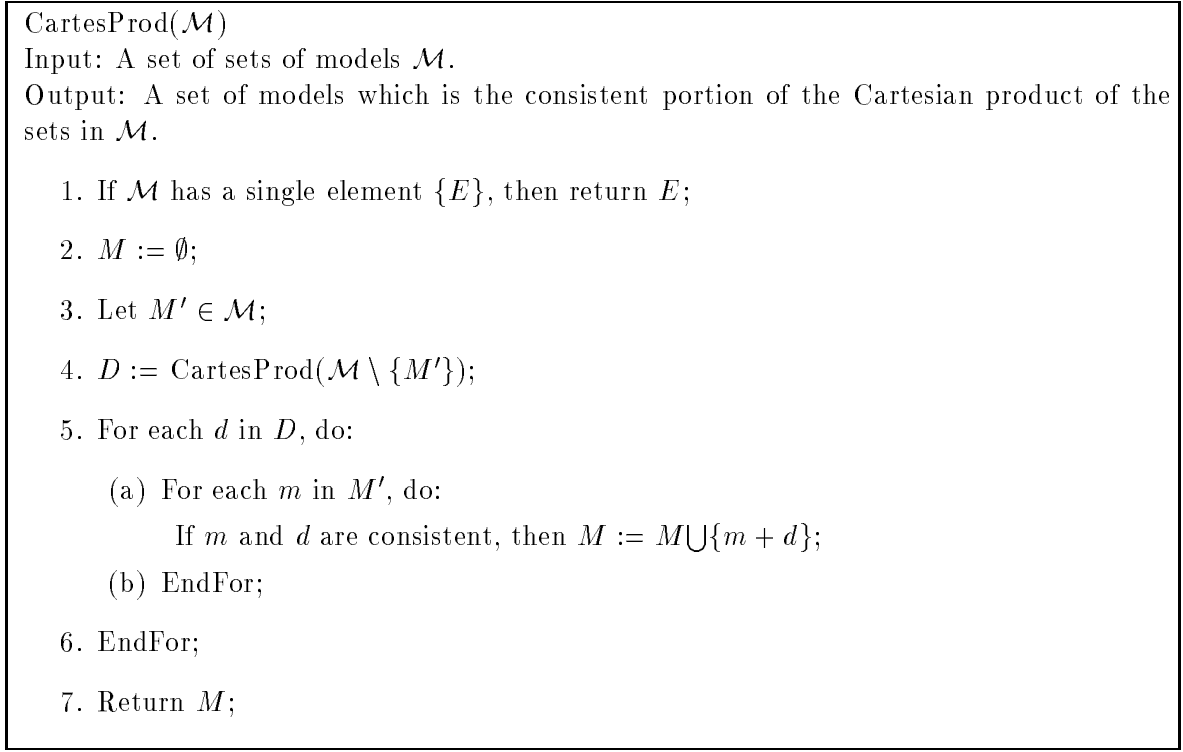

Figure 8: Procedure CartesProd

After executing step 1.d.i, we have $\Pi_{FM_m}$ set to

$$
\begin{aligned}
female \;\; &\longleftarrow \;\; not\ male \\
male \;\; &\longleftarrow \;\; not\ female
\end{aligned}
$$

The above knowledge base has two stable models: $\{female\}$ and $\{male\}$. The Cartesian product of the above set with $\{lion, mammal\}$ yields $M_{FM} = \{\{lion, mammal, female\},$ $\{lion, mammal, male\}\}$. At step 2, the Cartesian product of $M_{warm\_blooded}$, $M_{on\_land}$, and $M_{FM}$ is taken. Thus, the algorithm outputs $\{\{lion, mammal, on\_land, warm\_blooded, female\},$ $\{lion, mammal, on\_land, warm\_blooded, male\}\}$, and these are indeed the two stable models of $\Pi_0$. Note that algorithm AAS is more efficient than either ALL-STABLE1 or ALL-STABLE2 on the knowledge base $\Pi_0$.

**Theorem 4.2** *Algorithm AAS is correct, that is, $m$ is a stable model of a knowledge base $\Pi$ iff $m$ is generated by AAS when applied to $\Pi$.*

**Proof:** Let $s_0, s_1, ..., s_n$ be the ordering of the nodes of the super dependency graph by which the algorithm is executed. We can show by induction on $i$ that AAS, when at node $s_i$, generates all and only the stable models of the portion of the knowledge base composed of rules that only use atoms from $A_{s_i}$.

**case $i = 0$:** In this case, at step 1.d.ii of AAS, $\Pi_{s_m} = \Pi_s$; thus, the claim follows from the correctness of the algorithm ALL-STABLE called in step 1.d.ii.

**case $i > 0$:** Showing that every model generated is stable is straightforward, by the induction hypothesis and Theorem 2.2. The other direction is: suppose $m$ is a stable model of $\Pi_s$; show that $m$ is generated. Clearly, for each child $s$ of $s_i$, the projection of $m$ onto $A_s$ is a stable model of the part of the knowledge base that uses only atoms from $A_s$. By induction, $m_c$, which is the projection of $m$ onto the union of $A_s$ for every child $s$ of $s_i$, must belong to $M_{c(s_i)}$ computed at step 1.c. Therefore, to show that $m$ is generated, we need only show that $m' = m - m_c$ is a stable model of $\Pi_{s_{i_{m_c}}}$. This is easily done using Theorem 2.2.

$\square$

We will now analyze the complexity of AAS. First, given a knowledge base $\Pi$ and a set of atoms $s$, we define $\check{\Pi}_s$ to be the knowledge base obtained from $\Pi$ by deleting each negative occurrence of an atom that does not belong to $s$ from the body of every rule. For example, if $\Pi = \{a \longleftarrow not\ b, c \longleftarrow not\ d, a\}$ and $s = \{b\}$, then $\check{\Pi}_s = \{a \longleftarrow not\ b, c \longleftarrow a\}$. While visiting a node $s$ during the execution of AAS, we have to compute at step 1.d.ii all stable models of some knowledge base $\Pi_{s_m}$. Using either ALL-STABLE1 or ALL-STABLE2, the estimated time required to find all stable models of $\Pi_{s_m}$ is shorter than or equal to the time required to find all stable models of $\check{\Pi}_s$. This occurs because the number of negative atoms and the number of rules with negative atoms in their bodies in $\check{\Pi}_s$ is higher than or equal to the number of negative atoms and the number of rules with negative atoms in their bodies in $\Pi_{s_m}$, regardless of what $m$ is. Thus, if $\check{\Pi}_s$ is a Horn knowledge base, we can find the stable model of $\check{\Pi}_s$, and hence of $\Pi_{s_m}$, in polynomial time, no matter what $m$ is.

If $\hat{\Pi}_s$ is not positive, then we can find all stable models of $\hat{\Pi}_s$, and hence of $\Pi_{s_m}$, in time $min(ln * 2^k, ln * 2^c)$, where $l$ is the length of $\hat{\Pi}_s$, $n$ the number of atoms used in $\hat{\Pi}_s$, $c$ the number of rules in $\hat{\Pi}_s$ that contain negative atoms, and $k$ the number of atoms that appear negatively in $\hat{\Pi}_s$.

Then, with each knowledge base $\Pi$, we associate a number $t_\Pi$ as follows. Associate a number $v_s$ with every node in $G_\Pi$. If $\hat{\Pi}_s$ is a Horn knowledge base, then $v_s$ is 1; else, $v_s$ is $min(2^k, 2^c)$, where $c$ is the number of rules in $\hat{\Pi}_s$ that contain negative atoms from $s$, and $k$ is the number of atoms from $s$ that appear negatively in $\hat{\Pi}_s$. Now associate a number $t_s$ with every node $s$. If $s$ is a leaf node, then $t_s = v_s$. If $s$ has children $s_1, ..., s_j$ in $G_\Pi$, then $t_s = v_s * t_{s_1} * ... * t_{s_j}$. Define $t_\Pi$ to be $t_{s_1} * ... * t_{s_k}$, where $s_1, ..., s_k$ are all the sink nodes in $G_\Pi$.

**Definition 4.3** *A knowledge base $\Pi$ belongs to $\Omega_j$ if $t_\Pi = j$.*

**Theorem 4.4** *If a knowledge base belongs to $\Omega_j$ for some $j$, then it has at most $j$ stable models that can be computed in time $O(lnj)$.*

**Proof:** By induction on $j$. The dependency graph and the super dependency graph are both built in time linear in the size of the knowledge base. So we may only consider the time it takes to compute all stable models with the super dependency graph given.

**case $j = 1$:** $\Pi \in \Omega_1$ means that for every node $s$ in $G_\Pi$, $\hat{\Pi}_s$ is a Horn knowledge base. In other words, $\Pi$ is stratified, and therefore it has exactly one stable model. There are at most $n$ nodes in the graph. At each node, the loop in step 1.d is executed at most once, because at most one model is generated at every node. Procedure Convert runs in time $O(l_s)$, where $l_s$ is the length of $\Pi_s$ (we assume that $m$ is stored in an array where the access to each atom is in constant time). Since, for every node $s$, $\hat{\Pi}_s$ is a Horn knowledge base, $\Pi_{s_m}$ is computed in time $O(l_s n)$. Thus, the overall complexity is $O(ln)$.

**case $j > 1$:** By induction on $n$, the number of nodes in the super dependency graph of $\Pi$.

   **case $n = 1$:** Let $s$ be the single node in $G_\Pi$. Thus, $j = v_s$. Using the algorithms from Section 3, all stable models of $\Pi = \Pi_s$ can be found in time $O(lnv_s)$, and $\Pi$ has at most $v_s$ models.

   **case $n > 1$:** Assume without loss of generality that $G_\Pi$ has a single sink $s$ (to get a single sink, we can add to the program the rule $P \longleftarrow s_1, .., s_k$, where $s_1, ..., s_k$ are all the sinks and $P$ is a new atom). Let $c_1, ..., c_k$ be the children of $s$. For each child $c_i$, $\Pi(c_i)$, the part of the knowledge base which corresponds to the subgraph rooted by $c_i$, must belong to $\Omega_{t_i}$ for some $t_i \leq j$. By induction on $n$, for each child node $c_i$, all stable models of $\Pi(c_i)$ can be computed in time $O(lnt_i)$, and $\Pi(c_i)$ has at most $t_i$ stable models. Now let us observe what happens when AAS is visiting node $s$. First, the Cartesian product of all the models computed at the child nodes is taken. This is executed in time $O(n * t_1 * ... * t_k)$, and yields at most $t_1 * ... * t_k$ models in $M_{c(s)}$. For every $m \in M_{c(s)}$, we call Convert ($O(ln)$) and compute all the stable models of $\Pi_{s_m}$ ($O(lnv_s)$). We then combine them with $m$ using CartesProd ($O(nv_s)$). Thus, the overall complexity of computing $M_s$, that is, of computing all the stable models of $\Pi$, is $O(lnt_1 * ... * t_k * v_s) = O(lnj)$.

□

Note that all stratified knowledge bases belong to $\Omega_1$, and the more that any knowledge base looks stratified, the more efficient algorithm AAS will be.

Given a knowledge base $\Pi$, it is easy to find the minimum $j$ such that $\Pi$ belongs to $\Omega_j$. This follows because building $G_\Pi$ and finding $c$ and $k$ for every node in $G_\Pi$ are polynomial-time tasks. Hence,

**Theorem 4.5** *Given a knowledge base $\Pi$, we can find the minimum $j$ such that $\Pi$ belongs to $\Omega_j$ in polynomial time.*

**Example 4.6** For all the nodes $s$ in $G_{\Pi_0}$ except FM, $v_s=1$. $v_{FM} = 2$. Thus, $\Pi_0 \in \Omega_2$. $\Pi_1$ is a stratified knowledge base and therefore belongs to $\Omega_1$.

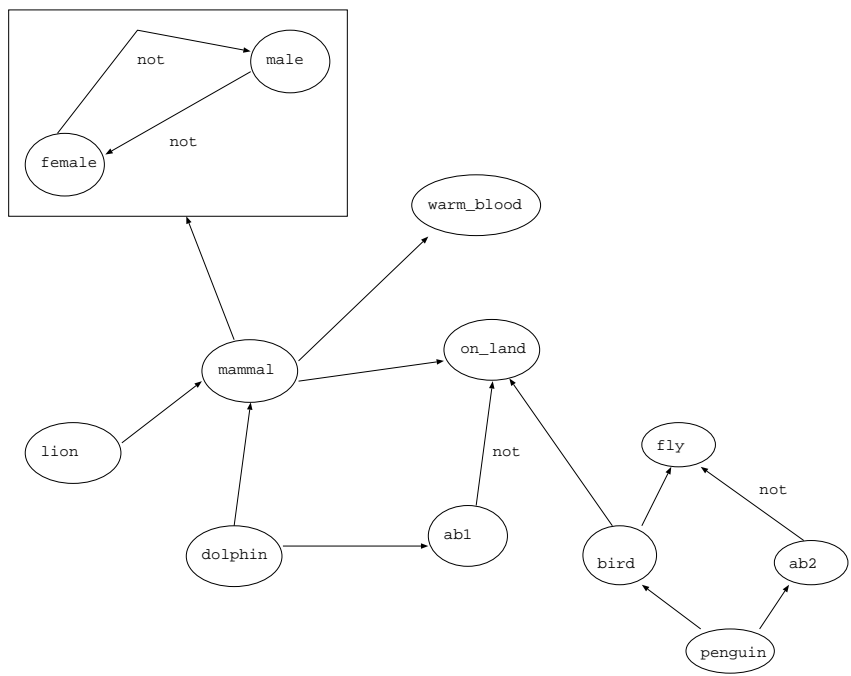

Figure 9: The super dependency graph of $\Pi_0 \bigcup \Pi_1$

The next example shows that step 5 of procedure CartesProd is necessary.

**Example 4.7** Consider knowledge base $\Pi_4$:

$$
\begin{array}{rcl}
a & \longleftarrow & not\, b \\
b & \longleftarrow & not\, a \\
c & \longleftarrow & a \\
d & \longleftarrow & b \\
e & \longleftarrow & c, d \\
f & \longleftarrow & c
\end{array}
$$

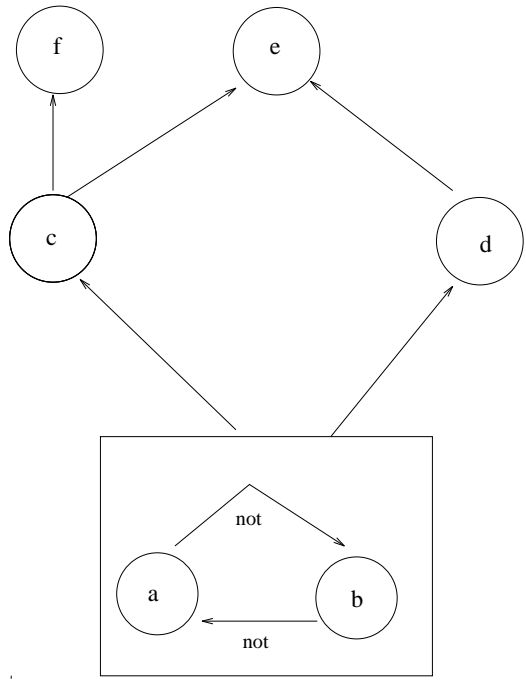

Figure 10: Super dependency graph of $\Pi_4$

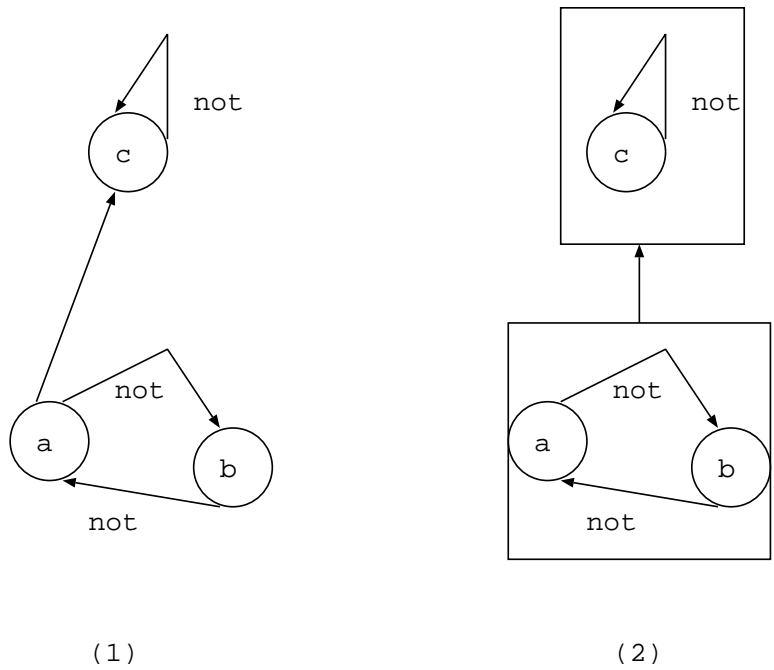

(1)                                    (2)

Figure 11: Dependency graph (1) and super dependency graph (2) of $\Pi_2$

The super dependency graph of $\Pi_4$ is shown in Figure 10. During the run of algorithm AAS, $M_{ab}$ (the set of models computed at the node $\{a, b\}$) is set to $\{\{a, \neg b\}, \{\neg a, b\}\}$. When AAS visits nodes $c$ and $d$, we get $M_c = \{\{a, \neg b, c\}, \{\neg a, b\}\}$, $M_d = \{\{\neg a, b, d\}, \{a, \neg b\}\}$. When AAS visits node $e$, CartesProd is called on the input $\{M_c, M_d\}$, yielding the output $M_e = \{\{a, \neg b, c\}, \{\neg a, b, d\}\}$. Note that CartesProd does not output any model in which both $c$ and $d$ are **true**, because the models $\{a, \neg b, c\}$ and $\{\neg a, b, d\}$ are inconsistent and CartesProd checks for consistency in step 5. When visiting node $f$, we get $M_f = \{\{a, \neg b, c, f\}, \{\neg a, b\}\}$. AAS then returns CartesProd($\{M_e, M_f\}$), which is $\{\{a, \neg b, c, f\}, \{\neg a, b, d\}\}$.

The next example demonstrates that some models generated at some nodes of the super dependency graph during the run of AAS may later be deleted, since they cannot be completed to a stable model of the whole knowledge base.

**Example 4.8** Consider knowledge base $\Pi_2$:

$$
\begin{aligned}
a &\longleftarrow \quad not\ b \\
b &\longleftarrow \quad not\ a \\
c &\longleftarrow \quad a, not\ c
\end{aligned}
$$

The dependency graph and the super dependency graph of $\Pi_2$ are shown in Figure 11. During the run of algorithm AAS, $M_{ab}$ (the set of models computed at the node $\{a, b\}$) is set to $\{\{a\}, \{b\}\}$. However, only $\{b\}$ is a stable model of $\Pi_2$.

Despite the deficiency illustrated in Example 4.8, algorithm AAS does have desirable features. First, AAS enables us to compute stable models in a modular fashion. We can use $G_\Pi$ as a structure in which to store the stable models. Once the knowledge base is changed, we need to resume computation only at the nodes affected by the change. For example, suppose that after computing the stable models of the knowledge base $\Pi_0$, we add to $\Pi_0$ the knowledge base $\Pi_1$ of Example 2.6, which gives us a new knowledge base, $\Pi_3 = \Pi_0 \bigcup \Pi_1$. The super dependency graph of the new knowledge base $\Pi_3$ is shown in Figure 9. Now we need only to compute the stable models at the nodes *penguin*, *bird*, *ab2*, *fly*, and *on_land* and then to combine the models generated at the sinks. We do not have to re-compute the stable models at all the other nodes as well.

Second, in using the AAS algorithm, we do not always have to compute all stable models up to the root node. If we are queried about an atom that is somewhere in the middle of the graph, it is often enough to compute only the models of the subgraph rooted by the node that represents this atom. For example, suppose we are given the knowledge base $\Pi_2$ and asked if *mammal* is **true** in every stable model of $\Pi_2$. We can run AAS for the nodes *dolphin*, *lion*, and *mammal* — and then stop. If *mammal* is **true** in all the stable models computed at the node *mammal* (i.e., in all the models in $M_{mammal}$), we answer "yes", otherwise, we must continue the computation.

Third, the AAS algorithm is useful in computing the labeling of a TMS subject to nogoods. A set of nodes of a TMS can be declared *nogood*, which means that all acceptable labeling should assign **false** to at least one node in the nogood set.[4] In stable models terminology, this means that when handling nogoods, we look for stable models in which at least one atom from a nogood is **false**. A straightforward approach would be to first compute all the stable models and then choose only the ones that comply with the nogood constraints. But since the AAS algorithm is modular and works from the bottom up, in many cases it can prevent the generation of unwanted stable models at an early stage. During the computation, we can exclude the submodels that do not comply with the nogood constraints and erase these submodels from $M_s$ once we are at a node $s$ in the super dependency graph such that $A_s$ includes all the members of a certain nogood.

---

4. In logic programming terminology nogoods are simply integrity constraints.

## 5. Computing Stable Models of First-Order Knowledge Bases

In this section, we show how we can generalize algorithm AAS so that it can find all stable models of a knowledge base over a first-order language with no function symbols. The new algorithm will be called FIRST-ACYCLIC-ALL-STABLE (FAAS).

We will now refer to a knowledge base as a set of rules of the form

$$C \longleftarrow A_1, A_2, ..., A_m, not\ B_1, ..., not\ B_n \tag{14}$$

where all $A$s, $B$s, and $C$ are atoms in a *first-order* language with no function symbols. The definitions of head, body, and positive and negative appearances of an atom are the same as in the propositional case. In the expression $p(X_1, ..., X_n)$, $p$ is called a *predicate name*.

As in the propositional case, every knowledge base $\Pi$ is associated with a directed graph called the *dependency graph* of $\Pi$, in which (a) each predicate name in $\Pi$ is a node, (b) there is a *positive arc* directed from a node $p$ to a node $q$ iff there is a rule in $\Pi$ in which $p$ is a predicate name in one of the $A_i$s and $q$ is a predicate name in the head, and (c) there is a *negative arc* directed from a node $p$ to a node $q$ iff there is a rule in $\Pi$ in which $p$ is a predicate name in one of the $B_i$s and $q$ is a predicate name in the head. The super dependency graph, $G_\Pi$, is defined in an analogous manner. We define a *stratified* knowledge base to be a knowledge base in which there are no cycles through the negative edges in the dependency graph of the knowledge base.

A knowledge base will be called *safe* iff each of its rules is safe. A rule is *safe* iff all the variables appearing in the head of the rule or in predicates appearing negative in the rule also appear in positive predicates in the body of the rule. In this section, we assume that knowledge bases are safe. The *Herbrand base* of a knowledge base is the set of all atoms constructed using predicate names and constants from the knowledge base. The set of *ground instances of a rule* is the set of rules obtained by consistently substituting variables from the rule with constants that appear in the knowledge base in all possible ways. The *ground instance of a knowledge base* is the union of all ground instances of its rules. Note that the ground instance of a first-order knowledge base can be viewed as a propositional knowledge base.

A *model* for a knowledge base is a subset $M$ of the knowledge base's Herbrand base. This subset has the property that for every rule in the grounded knowledge base, if all the atoms that appear positive in the body of the rule belong to $M$ and all the atoms that appear negative in the body of the rule do not belong to $M$, then the atom in the head of the rule belongs to $M$. A stable model for a first-order knowledge base $\Pi$ is a Herbrand model of $\Pi$, which is also a stable model of the grounded version of $\Pi$.

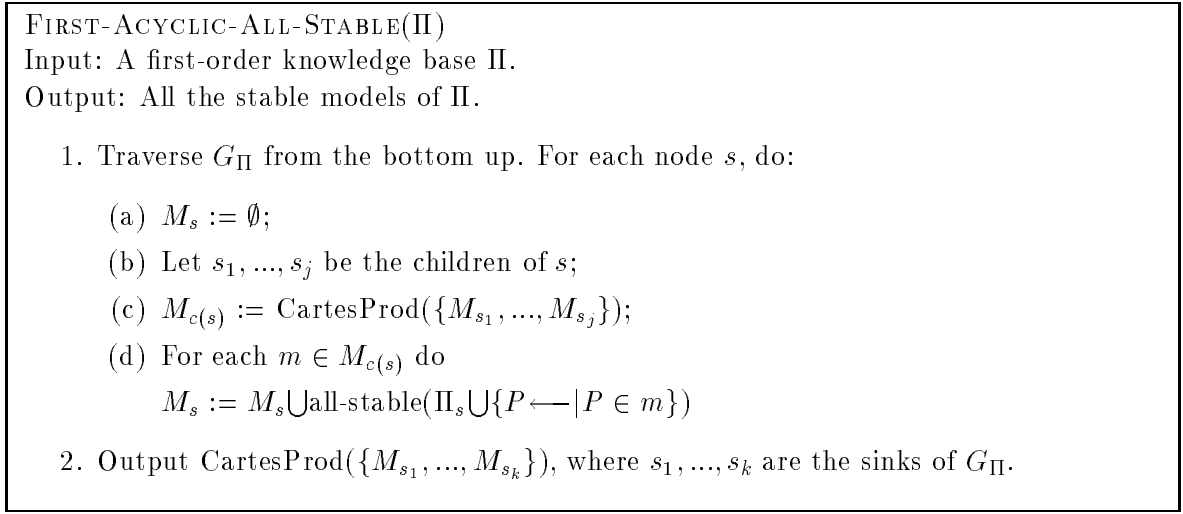

---

First-Acyclic-All-Stable(Π)
Input: A first-order knowledge base Π.
Output: All the stable models of Π.

1. Traverse $G_\Pi$ from the bottom up. For each node $s$, do:

   (a) $M_s := \emptyset$;

   (b) Let $s_1, ..., s_j$ be the children of $s$;

   (c) $M_{c(s)} := \mathrm{CartesProd}(\{M_{s_1}, ..., M_{s_j}\})$;

   (d) For each $m \in M_{c(s)}$ do
       $M_s := M_s \bigcup \mathrm{all\text{-}stable}(\Pi_s \bigcup \{P \longleftarrow \mid P \in m\})$

2. Output $\mathrm{CartesProd}(\{M_{s_1}, ..., M_{s_k}\})$, where $s_1, ..., s_k$ are the sinks of $G_\Pi$.

---

Figure 12: Algorithm First-Acyclic-All-Stable (FAAS)

We now present FAAS, an algorithm that computes all stable models of a first-order knowledge base. Let Π be a first-order knowledge base. As in the propositional case, with each node $s$ in $G_\Pi$ (the super dependency graph of Π), we associate $\Pi_s$, $A_s$, and $M_s$. $\Pi_s$ is the subset of Π containing all the rules about predicates whose names are in $s$. $A_s$ is the set of all predicate names $P$ that appear in the subgraph of $G_\Pi$ rooted by $s$. $M_s$ are the stable models associated with the sub–knowledge base of Π that contains only rules about predicates whose names are in $A_s$. Initially, $M_s$ is empty for every $s$. Algorithm FAAS traverses $G_\Pi$ from the bottom up. When at a node $s$, the algorithm first combines all the submodels of the children of $s$ into a single set of models, $M_{c(s)}$. Then, for each model $m$ in $M_{c(s)}$, it calls a procedure that finds all the stable models of $\Pi_s$ union the set of all unit clauses $P \longleftarrow$ where $P \in m$. The procedure All-Stable called by FAAS can be any procedure that computes all the stable models of a first-order knowledge base. Because procedure All-Stable computes stable models for only parts of the knowledge base, it may take advantage of some fractions of the knowledge base being stratified or having any other property that simplifies computation of the stable models of a fraction.

**Theorem 5.1** *Algorithm FAAS is correct, that is, $m$ is a stable model of a knowledge base Π iff $m$ is one of the models in the output when applying FAAS to Π.*

**Proof:** As the proof of Theorem 4.2. □

Note that the more that a knowledge base appears stratified, the more efficient algorithm FAAS becomes.

**Example 5.2** Consider knowledge base $\Pi_5$:

$$
\begin{aligned}
warm\_blooded(X) &\longleftarrow mammal(X) \\
live\_on\_land(X) &\longleftarrow mammal(X), not\, ab1(X) \\
female(X) &\longleftarrow mammal(X), not\, male(X)
\end{aligned}
$$

$$male(X) \longleftarrow mammal(X), not\, female(X)$$
$$mammal(X) \longleftarrow dolphin(X)$$
$$ab1(X) \longleftarrow dolphin(X)$$
$$mammal(X) \longleftarrow lion(X)$$
$$dolphin(flipper) \longleftarrow$$

$$live\_on\_land(X) \longleftarrow bird(X)$$
$$fly(X) \longleftarrow bird(X), not\, ab2(X)$$
$$bird(X) \longleftarrow penguin(X)$$
$$ab2(X) \longleftarrow penguin(X)$$
$$bird(bigbird) \longleftarrow$$

The super dependency graph of $\Pi_5$, $G_{\Pi_5}$, is the same as the super dependency graph of the knowledge base $\Pi_2$ (see Figure 9). Observe that when at node *mammal*, for example, in step 1.d the algorithm looks for all stable models of the knowledge base $\Pi' = \Pi_{mammal} \bigcup \{\longleftarrow dolphin(flipper)\}$, where $\Pi_{mammal} = \{mammal(X) \longleftarrow dolphin(X), mammal(X) \longleftarrow lion(X)\}$. $\Pi'$ is a stratified knowledge base that has a unique stable model that can be found efficiently. Hence, algorithm FAAS saves us from having to ground all the rules of the knowledge base before starting to calculate the models, and it can take advantage of parts of the knowledge base being stratified.

## 6. Related Work

In recent years, quite a few algorithms have been developed for reasoning with stable models. Nonetheless, as far as we know, the work presented here is original in the sense that it provides a partition of the set of all the knowledge bases into a hierarchy of tractable classes. The partition is based on the structure of the dependency graph. Intuitively, the task of computing all the stable models of a knowledge base using algorithm AAS becomes increasingly complex as the "distance" of the knowledge base from being stratified becomes larger. Next, we summarize the work that seems to us most relevant.

Algorithm AAS is based on an idea that appears in the work of Lifschitz and Turner (1994), where they show that in many cases a logic program can be divided into two parts, such that one part, the "bottom" part, does not refer to the predicates defined in the "top" part. They then explain how the task of computing the stable models of a program can be simplified when the program is split into parts. Algorithm AAS, using the superstructure of the dependency graph, exploits a specific method for splitting the program.

Bell et al. (1994) and Subrahmanian et al. (1995) implement linear and integer programming techniques in order to compute stable models (among other nonmonotonic logics). However, it is difficult to assess the merits of their approaches in terms of complexity. Ben-Eliyahu and Dechter (1991) illustrate how a knowledge base $\Pi$ can be translated into a propositional theory $T_\Pi$ such that each model of the latter corresponds to a stable model of the former. It follows from this that the problem of finding all the stable models of a knowledge base corresponds to the problem of finding all the models of a propositional theory. Satoh and Iwayama (1991) provide a nondeterministic procedure for computing the stable models of logic programs with integrity constraints. Junker and Konolige (1990) present an algorithm for computing TMS' labels. Antoniou and Langetepe (1994) introduce a method for representing some classes of default theories as normal logic programs in such a way that SLDNF-resolution can be used to compute extensions. Pimentel and Cuadrado (1989) develop a label-propagation algorithm that uses data structures called *compressible semantic trees* in order to implement a TMS; their algorithm is based on stable model semantics. The algorithms developed by Marek and Truszczyński (1993) for autoepistemic logic can also be adopted for computing stable models. The procedures by Marek and Truszczyński (1993), Antoniou and Langetepe (1994), Pimentel and Cuadrado (1989), Ben-Eliyahu and Dechter (1991), Satoh and Iwayama (1991), Bell et al. (1994), Subrahmanian et al. (1995), and Junker and Konolige (1990) do not take advantage of the structure of the knowledge base as reflected in its dependency graph, and therefore are not efficient for stratified knowledge bases.

Saccà and Zaniolo (1990) present a *backtracking fixpoint* algorithm for constructing *one* stable model of a first-order knowledge base. This algorithm is similar to algorithm ALL-STABLE2 presented here in Section 3 but its complexity is worse than the complexity of ALL-STABLE2. They show how the backtracking fixpoint algorithm can be modified to handle stratified knowledge bases in an efficient manner, but the algorithm needs further adjustments before it can deal efficiently with knowledge bases that are very close to being stratified. Leone et al. (1993) present an improved backtracking fixpoint algorithm for computing one stable model of a Datalog¬ program and discuss how the improved algorithm can be implemented. One of the procedures called by the improved algorithm is based on the backtracking fixpoint algorithm of Saccà and Zaniolo (1990). Like the backtracking fixpoint algorithm, the improved algorithm as is does not take advantage of the structure of the program, i.e., it is not efficient for programs that are close to being stratified.

Several tractable subclasses for computing extensions of default theories (and, hence, computing stable models) are known (Kautz & Selman, 1991; Papadimitriou & Sideri, 1994; Palopoli & Zaniolo, 1996; Dimopoulos & Magirou, 1994; Ben-Eliyahu & Dechter, 1996). Some of these tractable subclasses are characterized using a graph that reflects dependencies in the program between atoms and rules. The algorithms presented in these papers are complete only for a subclass of all knowledge bases, however. Algorithms for computing extensions of *stratified* default theories or extensions of default theories that have no odd cycles (in some precise sense) are given by Papadimitriou and Sideri (1994) and Cholewiński (1995a, 1995b).

Algorithms for handling a TMS with nogoods have been developed in the AI community by Doyle (1979) and Charniak et al. (1980). But, as Elkan (1990) points out, these algorithms are not always faithful to the semantics of the TMS and their complexities have not been analyzed. Dechter and Dechter (1994) provide algorithms for manipulating a TMS when it is represented as a constraint network. The efficiency of their algorithms depends on the structure of the constraint network representing the TMS, and the structure they employ differs from the dependency graph of the knowledge base.

## 7. Conclusion

The task of computing stable models is at the heart of several systems central to AI, including TMSs, autoepistemic logic, and default logic. This task has been shown to be NP-hard. In this paper, we present a partition of the set of all knowledge bases to classes $\Omega_1, \Omega_2, ...$, such that if a knowledge base $\Pi$ is in $\Omega_k$, then $\Pi$ has at most $k$ stable models, and they may all be found in time $O(lnk)$, where $l$ is the length of the knowledge base and $n$ the number of atoms in $\Pi$. Moreover, for an arbitrary knowledge base $\Pi$, we can find the minimum $k$ such that $\Pi$ belongs to $\Omega_k$ in time linear in the size of $\Pi$. Intuitively, the more the knowledge base is stratified, the more efficient our algorithm becomes. We believe that beyond stratified knowledge bases, the more expressive the knowledge base is (i.e. the more rules with nonstratified negation in the knowledge base), the less likely it will be needed. Hence, our analysis should be quite useful. In addition, we show that algorithm AAS has several advantages in a dynamically changing knowledge base, and we provide applications for answering queries and implementing a TMS's nogood strategies. We also illustrate a generalization of algorithm AAS for the class of first-order knowledge bases.

Algorithm AAS can easily be adjusted to find only one stable model of a knowledge base. While traversing the super dependency graph, we generate only one model at each node. If we arrive at a node where we cannot generate a model based on what we have computed so far, we backtrack to the most recent node where several models were available to choose from and take the next model that was not yet chosen. The worst-case time complexity of this algorithm is equal to the worst-case time complexity of the algorithm for finding all stable models because we may have to exhaust all possible ways of generating a stable model before finding out that a certain knowledge base does not have a stable model at all. Nevertheless, we believe that in the average case, finding just one model will be easier than finding them all. A similar modification of the AAS algorithm is required if we are interested in finding one model in which one particular atom gets the value **true**.

This work is another attempt to bridge the gap between the declarative systems (e.g., default logic, autoepistemic logic) and the procedural systems (e.g., ATMs, Prolog) of the nonmonotonic reasoning community. It is argued that while the declarative methods are sound, they are impractical since they are computationally expensive, and while the procedural methods are more efficient, it is difficult to completely understand their performance or to evaluate their correctness. The work presented here illustrates that the declarative and the procedural approaches can be combined to yield an efficient yet formally supported nonmonotonic system.

## Acknowledgments

Thanks to Luigi Palopoli for useful comments on earlier draft of this paper and to Michelle Bonnice and Gadi Dechter for editing on parts of the manuscript. Many thanks to the anonymous referees for very useful comments.

Some of this work was done while the author was visiting the Cognitive Systems Laboratory, Computer Science Department, University of California, Los Angeles, California, USA. This work was partially supported by NSF grant IRI-9420306 and by Air Force Office of Scientific Research grant #F49620-94-1-0173.

# References

Antoniou, G., & Langetepe, E. (1994). Soundness and completeness of a logic programming approach to default logic. In *AAAI-94: Proceedings of the 12th national conference on artificial intelligence*, pp. 934–939. AAAI Press, Menlo Park, Calif.

Apt, K., Blair, H., & Walker, A. (1988). Towards a theory of declarative knowledge. In Minker, J. (Ed.), *Foundations of deductive databases and logic programs*, pp. 89–148. Morgan Kaufmann.

Bell, C., Nerode, A., Ng, R., & Subrahmanian, V. (1994). Mixed integer programming methods for computing non-monotonic deductive databases. *Journal of the ACM*, *41*(6), 1178–1215.

Ben-Eliyahu, R., & Dechter, R. (1994). Propositional semantics for disjunctive logic programs. *Annals of Mathematics and Artificial Intelligence*, *12*, 53–87. A short version appears in JICSLP-92: Proceedings of the 1992 joint international conference and symposium on logic programming.

Ben-Eliyahu, R., & Dechter, R. (1996). Default reasoning using classical logic. *Artificial Intelligence*, *84*(1-2), 113–150.

Bidoit, N., & Froidevaux, C. (1987). Minimalism subsumes default logic and circumscription in stratified logic programming. In *LICS-87: Proceedings of the IEEE symposium on logic in computer science*, pp. 89–97. IEEE Computer Science Press, Los Alamitos, Calif.

Charniak, E., Riesbeck, C. K., & McDermott, D. V. (1980). *Artificial Intelligence Programming*, chap. 16. Lawrence Erlbaum, Hillsdale, NJ.

Cholewiński, P. (1995a). Reasoning with stratified default theories. In Marek, W. V., Nerode, A., & Truszczyński, M. (Eds.), *Logic programming and nonmonotonic reasoning: proceedings of the 3rd international conference*, pp. 273–286. Lecture notes in computer science, 928. Springer-Verlag, Berlin.

Cholewiński, P. (1995b). Stratified default theories. In Pacholski, L., & Tiuryn, A. (Eds.), *Computer science logic: 8th workshop, CSL'94: Selected papers*, pp. 456–470. Lecture notes in computer science, 933. Springer-Verlag, Berlin.

Dechter, R., & Dechter, A. (1996). Structure-driven algorithms for truth maintenance. *Artificial Intelligence*, *82*(1-2), 1–20.

Dimopoulos, Y., & Magirou, V. (1994). A graph-theoretic approach to default logic. *Journal of Information and Computation*, *112*, 239–256.

Doyle, J. (1979). A truth-maintenance system. *Artificial Intelligence*, *12*, 231–272.

Dung, P. M. (1991). Negation as hypothesis: An abductive foundation for logic programming. In Furukawa, K. (Ed.), *ICLP-91: Proceedings of the 8th international conference on logic programming*, pp. 3–17. MIT Press.

Elkan, C. (1990). A rational reconstruction of nonmonotonic truth maintenance systems. *Artificial Intelligence*, *43*, 219–234.

Eshghi, K., & Kowalski, R. A. (1989). Abduction compared with negation by failure. In Levi, G., & Martelli, M. (Eds.), *ICLP-89: Proceedings of the 6th international conference on logic programming*, pp. 234–254. MIT Press.

Fine, K. (1989). The justification of negation as failure. *Logic, Methodology and Philosophy of Science*, *8*, 263–301.

Gelfond, M. (1987). On stratified autoepistemic theories. In *AAAI-87: Proceedings of the 5th national conference on artificial intelligence*, pp. 207–211. Morgan Kaufmann.

Gelfond, M., & Lifschitz, V. (1988). The stable model semantics for logic programming. In Kowalski, R. A., & Bowen, K. A. (Eds.), *Logic Programming: Proceedings of the 5th international conference*, pp. 1070–1080. MIT Press.

Gelfond, M., & Lifschitz, V. (1991). Classical negation in logic programs and disjunctive databases. *New Generation Computing*, *9*, 365–385.

Junker, U., & Konolige, K. (1990). Computing the extensions of autoepistemic and default logics with a TMS. In *AAAI-90: Proceedings of the 8th national conference on artificial intelligence*, pp. 278–283. AAAI Press.

Kakas, A. C., & Mancarella, P. (1991). Stable theories for logic programs. In Saraswat, V., & Udea, K. (Eds.), *ISLP-91: Proceedings of the 1991 international symposium on logic programming*, pp. 85–100. MIT Press.

Kautz, H. A., & Selman, B. (1991). Hard problems for simple default logics. *Artificial Intelligence*, *49*, 243–279.

Leone, N., Romeo, N., Rullo, M., & Saccà, D. (1993). Effective implementation of negation in database logic query languages. In Atzeni, P. (Ed.), *LOGIDATA+: Deductive database with complex objects*, pp. 159–175. Lecture notes in computer science, 701. Springer-Verlag, Berlin.

Lifschitz, V., & Turner, H. (1994). Splitting a logic program. In Van Hentenryck, P. (Ed.), *ICLP-94: Proceedings of the 11th international conference on logic programming*, pp. 23–37. MIT Press.

Marek, V. W., & Truszczyński, M. (1993). *Nonmonotonic logic: Context-dependent reasoning*. Springer Verlag, Berlin.

Marek, W., & Truszczyński, M. (1991). Autoepistemic logic. *Journal of the ACM*, *38*, 588–619.

Moore, R. C. (1985). Semantical consideration on nonmonotonic logic. *Artificial Intelligence*, *25*, 75–94.

Palopoli, L., & Zaniolo, C. (1996). Polynomial-time computable stable models.. *Annals of Mathematics and Artificial Intelligence*, in press.

Papadimitriou, C. H., & Sideri, M. (1994). Default theories that always have extensions. *Artificial Intelligence*, *69*, 347–357.

Pimentel, S. G., & Cuadrado, J. L. (1989). A truth maintenance system based on stable models. In Lusk, E. L., & Overbeek, R. A. (Eds.), *ICLP-89: Proceedings of the 1989 North American conference on logic programming*, pp. 274–290. MIT Press.

Przymusinska, H., & Przymusinski, T. (1990). Semantic issues in deductive databases and logic programs. In Banerji, R. B. (Ed.), *Formal techniques in artificial intelligence: A sourcebook*, pp. 321–367. North-Holland, New York.

Reiter, R. (1980). A logic for default reasoning. *Artificial Intelligence*, *13*, 81–132.

Saccà, D., & Zaniolo, C. (1990). Stable models and non-determinism in logic programs with negation. In *PODS-90: Proceedings of the 9th ACM SIGACT-SIGMOD-SIGART symposium on principles of database systems*, pp. 205–217. ACM Press.

Satoh, K., & Iwayama, N. (1991). Computing abduction by using the TMS. In Furukawa, K. (Ed.), *ICLP-91: Proceedings of the 8th international conference on logic programming*, pp. 505–518. MIT Press.

Subrahmanian, V., Nau, D., & Vago, C. (1995). WFS + branch and bound = stable models. *IEEE Transactions on Knowledge and Data Engineering*, *7*(3), 362–377.

Tarjan, R. (1972). Depth-first search and linear graph algorithms. *SIAM Journal on Computing*, *1*, 146–160.